\documentclass[default,iicol]{sn-jnl}

\usepackage{graphicx}%
\usepackage{multirow}%
\usepackage{amsmath,amssymb,amsfonts}%
\usepackage{amsthm}%
\usepackage{mathrsfs}%
\usepackage{xcolor}%
\usepackage{textcomp}%
\usepackage{manyfoot}%
\usepackage{booktabs}%
\usepackage{algorithm}%
\usepackage{algorithmicx}%
\usepackage{algpseudocode}%
\usepackage{listings}%
\usepackage{todonotes}
\usepackage{tabularx}
\usepackage{hyperref}

\usepackage{enumerate}
\usepackage[shortlabels]{enumitem}
\usepackage{graphicx}
\usepackage{subcaption}

\newcommand{\etal}[0]{\textit{et al.}}

\begin{document}

\title[Enhancing Robot Assistive Behaviour with Reinforcement Learning and Theory of Mind]{Enhancing Robot Assistive Behaviour with Reinforcement Learning and Theory of Mind}


\author*[1]{\fnm{Antonio} \sur{Andriella}}\email{antonio.andriella@iiia.csic.es}

\author[2]{\fnm{Giovanni} \sur{Falcone}}\email{giova.falcone@studenti.unina.it}

\author*[2]{\fnm{Silvia} \sur{Rossi}}\email{silvia.rossi@unina.it}

\affil*[1]{\orgname{Artificial Intelligence Research Institute (IIIA-CSIC)}, \orgaddress{\city{Barcelona}, \postcode{08193}, \country{Spain}}}

\affil[2]{\orgdiv{Department of Electrical Engineering and Information Technologies}, \orgname{University of Naples Federico II}, \orgaddress{\city{Naples}, \postcode{08005},  \country{Italy}}}


\abstract{The adaptation to users' preferences and the ability to infer and interpret humans' beliefs and intents, which is known as the Theory of Mind (ToM), are two crucial aspects for achieving effective human-robot collaboration.
Despite its importance, very few studies have investigated the impact of adaptive robots with ToM abilities.
In this work, we present an exploratory comparative study to investigate how social robots equipped with ToM abilities impact user's performance and perception.
We design a two-layer architecture. The Q-learning agent on the first layer learns the robot's higher-level behaviour. On the second layer, a heuristic-based ToM infers the user's intended strategy and is responsible for implementing the robot's assistance, as well as providing the motivation behind its choice.
We conducted a user study in a real-world setting, involving 56 participants who interacted with either an adaptive robot capable of ToM, or with a robot lacking such abilities. Our findings suggest that participants in the ToM condition performed better, accepted the robot's assistance more often, and perceived its ability to adapt, predict and recognise their intents to a higher degree. Our preliminary insights could inform future research and pave the way for designing more complex computation architectures for adaptive behaviour with ToM capabilities.}


\maketitle

\section{Introduction}
\label{sec:introduction}
Cognitive stimulation is crucial for maintaining and improving abilities such as memory, attention, and executive function~\cite{Woods_CDSR23}. Regular engagement in activities that challenge and stimulate the brain has been shown to positively impact cognitive health and can delay the onset of age-related declines~\cite{Kelly_ARR14}. The use of socially assistive robots (SARs) in memory exercises~\cite{Alrazaq_JMIR22, Ning_IEEETEHM20} has the potential to provide a personalised and engaging platform for delivering cognitive stimulation, offering a unique and interactive experience for users~\cite{Maggi_IJSR21}.
Indeed, robots have been proven to be very effective in performing simple, and repetitive tasks, making them a perfect tool to support healthcare professionals and enhance their effectiveness during their daily working routine~\cite{Tapus_ieeerr09}. Nonetheless, for robots to be most effective in providing assistance, they must cater to the specific needs of users and aim to prevent negative emotions such as frustration (due to overly difficult exercises) or boredom (due to overly simple exercises) during cognitive exercises. It is important that the robot's approach aligns precisely with the user's requirements to achieve optimal results. 

In recent years, several studies have demonstrated the impact of robot adaptivity on users' performance as well as on their engagement in assistive tasks ~\cite{Andriella_CCJ19, Tsiakas_T18, Hemminghaus_hri17}. 
However, adaptation in the robot's behaviour may result in affecting people's ability to understand and predict the robot, impacting their trust. To avoid this, it has to be noted that humans are more likely to cooperate with machines with whom they can share a mental representation. Hence, the robot should be able to have a Theory of Mind (ToM) of the users over time and personalise its behaviour
according to the inferred beliefs and intentions~\cite{Scasellati_AR02, Shvo_iros22, Soderlund_JRCS22}.
Despite those works, very little is known about how to effectively combine adaptivity with ToM to improve task performance and increase the user's perceived competence of the robot~\cite{Bianco_hri20}. 
Therefore, in this work, we aim to fill this research gap.
We build upon our previous knowledge, wherein we demonstrated how a robot can tailor its degrees of assistance to patients affected by cognitive decline~\cite{Andriella_UMUAI22}. Here, we go a step further by endowing the robot with the capability to understand users' strategies through a process of mentalising and, therefore, use that information to provide qualitatively better assistance.
This study takes an exploratory approach to address the following research question through a comparative study: \textit{would a robot endowed with adaptive socially assistive behaviour and ToM abilities have a different impact on users’ performance and perceived robot capabilities compared to an adaptive robot without ToM abilities?}

\begin{figure}[t!]
        \centering    	
        \includegraphics[width=\linewidth]{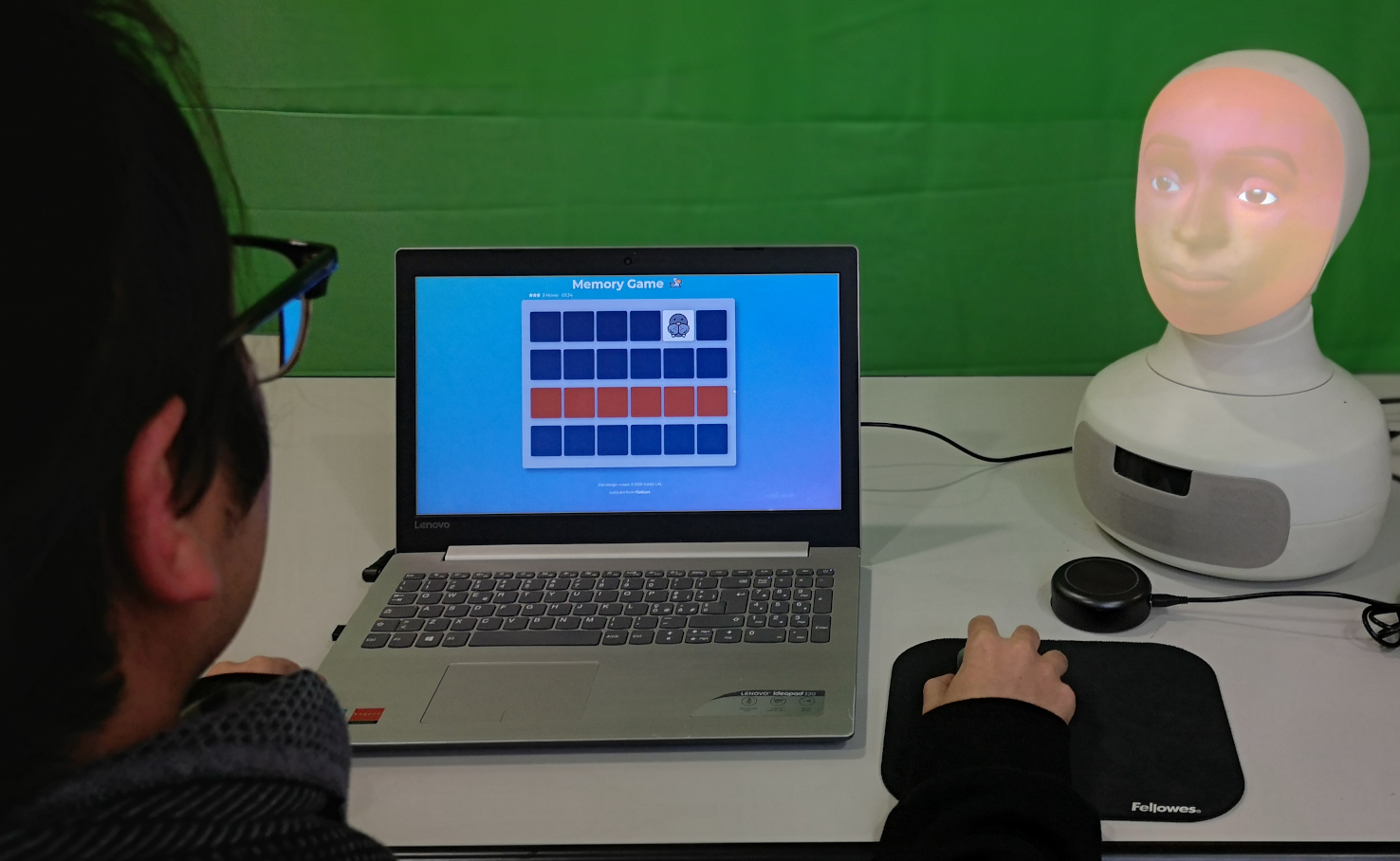}
    	\caption{A user playing the memory game with the assistance of the Furhat robot.}    	\label{fig:user_playing}
\end{figure} 

To tackle this research question, we propose a computational approach in which a robot learns the socially assistive behaviour that best fits the users while providing advice that relies on the user's beliefs and their intended strategies to solve a memory game (see Figure~\ref{fig:user_playing}).
Specifically, we create a two-layer architecture. The upper level, namely the learning layer, includes a Q-learning-based agent trained in simulation to model an imperfect player and learn the policy that best fits the user's needs (e.g., suggest the card). The lower level, namely the mentalising layer, consists of a heuristic-based ToM that based on the previous history of users' moves is used to estimate their strategies and their beliefs about the card's selection. Here, ToM has employed both for operationalising the assistance provided by the RL (e.g.,  suggest shark) and explain the rationale behind its hint (``\textit{You have seen the shark several times, the other card is in row 1 col 2, you should remember the location}").
To evaluate our system, we carried out a user study in a real-world setting. N=56 untrained participants played a memory game with the assistance of the Furhat robot during a national fair. 
We found that participants who were assisted by a robot capable of ToM performed better, accepted the hints provided by the robot more frequently, and perceived the robot as more capable of adapting, predicting, and recognising their intentions in comparison to those participants who interacted with a robot without ToM abilities. 

The findings of this study provide insights that can inform future research into the design of robots that exhibit adaptive behaviour tailored to users' capacities, while also incorporating ToM capabilities. 

In summary, the contributions of our study are the following:

\begin{itemize}
    \item Development of a hierarchical architecture that learns: i) socially assistive actions in simulation along, with ii) user's intended strategies in real interactions,
    \item Deployment and evaluation of a social robot endowed with such architecture in a real-world setting with 56 untrained participants.
\end{itemize}

\section{Related Work}
\label{sec:related_work}
In the context of designing robots intended to interact with diverse users, adaptivity emerges as a key requirement. Human performance in completing tasks can vary significantly due to a multitude of factors, including individual preferences, cognitive capabilities, and the inherent complexity of the task. In pursuit of enhancing robot adaptivity for humans and mitigating possible unpredictability in the behaviour, researchers are starting to focus on endowing robots with ToM capabilities. The integration of ToM into robots offers a dual advantage. Firstly, it can enhance human-robot interactions by tailoring the assistance provided to users based on a deeper understanding of their needs. Secondly, it can foster trust and transparency in the robot's actions as it becomes capable of providing explanations for its decisions.

This section delves into the domain of robot adaptivity and explores the most relevant contributions to the topic (see Section~\ref{sec:robot_adaptive_behaviour}). Additionally, it provides a comprehensive overview of relevant studies that have investigated the benefits of endowing robots with ToM, first exploring some related concepts such as intention recognition, transparency, and explainability, and then also ToM approaches that concentrated on developing computational models to simulate this cognitive capability (see Section~\ref{sec:robot_theory_of_mind}).

    \subsection{Robot Adaptive Behaviour}
    \label{sec:robot_adaptive_behaviour}
    Previous work on robot adaptivity has shown the importance of tailoring the robot's behaviour to humans in assistive tasks.
    
    Interesting contributions have emerged in the educational context. Park~\etal\cite{Park_aaai19} employed RL to determine which stories to select in order to optimise children's engagement and enhance their linguistic skills during storytelling activities. Senft~\etal\cite{Senft_SR19} introduced SPARC, a novel framework aimed at addressing the sparsity of reward functions in RL. This approach involved a human ``wizard'' who initially guided the robot's learning process by selecting actions. Over time, control gradually transitioned to the robot.
    A different approach from RL emerged in the field of intelligent tutoring systems (ITS), where the system adapts to the learner by building dynamic models of their knowledge, skills, and learning preferences. These models are constructed by continuously monitoring and analysing learner interactions, as noted by Zhang~\etal\cite{Zhang_IEEETLT24}. Expanding on this idea, Leyzberg~\etal\cite{Leyzberg_THRI18} explored the role of personalised lesson sequencing in robot tutoring through an adaptive Hidden Markov Model (HMM) that tracked student proficiency.  
    Schodde~\etal\cite{Schodde_hri17} took a complementary approach, utilising an extended Bayesian Knowledge Tracing (BKT) model combined with predictive decision-making to dynamically adjust tutoring strategies. By tracking the learner’s knowledge state, their robot tutor adapted its next steps accordingly, demonstrating improved learning outcomes in an L2 learning game compared to a random control group. Moving beyond cognitive modelling, Spaulding~\etal\cite{Spaulding_amas16} integrated affective data into BKT models, focusing on how emotional signals like engagement and frustration can improve knowledge inference. Their study showed that children interacting with an affect-aware social robot displayed stronger engagement, indicating the potential of affective data to enhance educational experiences. Building on these developments, Salomons~\etal\cite{Salomons_FRAI24} introduced Time-Dependent Bayesian Knowledge Tracing (TD-BKT) to track skill acquisition in complex tasks. By incorporating time-sensitive parameters and accounting for incomplete data, TD-BKT further refined skill tracking and demonstrated improved accuracy, especially in tasks like electronic circuit building, where traditional models struggled. 

    In the healthcare context, robots have been designed to adapt to the patient's needs to provide personalized assistance, enhance engagement, and support therapeutic goals. Raggioli~\etal\cite{Raggioli} proposed a reinforcement learning (RL) approach that allowed the robot to adaptively determine the optimal monitoring distance and direction based on the user's current activity, which was estimated using data from wearable devices. Maroto~\etal\cite{Maroto_CIS23} presented a decision-making system that, leveraging user information and a biologically-inspired module, personalised interactions to maintain user engagement over time. Moro~\etal\cite{Moro_THRI18} explored the development of a novel learning architecture for socially assistive robots, particularly focusing on aiding individuals with cognitive impairments. The authors argue that for these robots to be effective, they must not only display appropriate behaviours but also personalise these behaviours according to the user's cognitive abilities. Their work combined Learning from Demonstration (LfD) and  RL to enable robots to learn and adapt their behaviours based on user interactions.    
    
    The importance of adaptivity has also been explored in the specific context of cognitive training exercises to which this work aims.
    Chan~\etal\cite{Chan_IJARS12} proposed a control architecture based on hierarchical RL that enabled a robot to offer adaptive assistance to humans. Their findings showed that individuals performed better in a memory task when the robot provided support that matched participants' needs. 
    Similarly, Tsiakas~\etal \cite{Tsiakas_T18} introduced an Interactive RL system that allowed a Nao robot to offer adaptive support using the user's task performance and engagement. More recently, Umbrico~\etal\cite{Umbrico_IJSR23} proposed MIRIAM, a cognitive architecture inspired by the dual process theory. MIRIAM consisted of two reasoning layers: a faster, reactive layer based on machine learning and a slower, deliberative layer based on planning. By combining these layers, the system could provide personalized support. 
    In our previous work~\cite{Andriella_UMUAI22}, we demonstrated the importance of taking individual differences into account when designing adaptive support systems for cognitive training. We proposed CARESSER, a framework that used therapists' demonstrations and expertise to learn the assistance that best fit the patient's unique needs in cognitive therapy with older adults affected by mild cognitive impairment. 
    
    Despite the insightful results, these studies either incorporated the user's model into the decision-making process, or they did not explicitly account for the user's belief. In this work, we develop a hierarchical computation architecture in which we decouple the learning of the robot's assistive behaviours from their execution. In this way, the learning layer is responsible for selecting the assistive action while the mentalising layer is accountable for implementing that action according to the user's current inferred strategy during the game, so making the decision more transparent to the user.
    
    \subsection{Theory of Mind}
    \label{sec:robot_theory_of_mind}
    ToM is critical in enabling robots to infer and reason about the mental states of others, including beliefs, desires, and intentions. By equipping robots with ToM capabilities, they can provide more personalised and context-aware assistance, particularly in assistive tasks.
    
    Intention recognition plays a pivotal role in realising ToM capabilities, as it allows robots to predict user goals based on their actions, behaviours, and contextual cues. 
    Mavsar~\etal\cite{Mavsar_icar21} introduced a method using recurrent neural networks (OptiNet and HandNet) to predict human intentions in dynamic human-robot collaboration.
    Similarly, Chang~\etal\cite{Chang_iros18} explored how robots can combine intent recognition with communication through legible motions to improve task performance. 
    A very interesting work in the field was presented by Jain~\etal\cite{Jain_THRI20} who introduced a formal mathematical formulation for intent inference during assistive teleoperation under shared autonomy. They introduced a Bayesian filtering technique for probabilistic reasoning about user goals in shared autonomy. The robot in their system infers intentions based on multiple non-verbal cues, adapting dynamically to assist in real-time. In our work, we do not utilise any advanced AI learning techniques; instead, we estimate the user's intended course of action through a heuristic-based approach grounded in the specific logic of the assistive task used in the study.
    
    On the other hand, system transparency and explainability are critical in ensuring users understand and trust the robot's behaviour. In prior research, system transparency and the provision of explanations have been shown to significantly affect user perceptions of trust, satisfaction, and usability~\cite{Sakai_AR22}. For instance, Kizilcec~\etal\cite{Kizilcec_chi16} found that transparency in decision-making systems improved users' trust and understanding of the system, even when the system made errors. Similarly, Wang~\etal\cite{Wang_chi19} demonstrated that explanations that align with a user's mental model can enhance task performance and satisfaction. In human-robot interaction, Ezenyilimba~\etal\cite{Ezenyilimba_JCEDM23} showed that providing clear, human-centred explanations improves user trust and situation awareness in human-robot teams. Olivares~\etal\cite{Olivares_icra23} evaluated the impact of explanation specificity using an ontology in an online study in the context of collaborative robotics.  
    Angelopoulos~\etal\cite{Angelopoulos_icsr23} showed how integrating explanatory mechanisms into robot actions can enhance transparency and mutual understanding in an online user study. Interestingly enough, the level of transparency was found to be correlated to participants' existing knowledge. Similarly, Nair~\etal\cite{Nair_icsr23} found that providing brief explanations increased transparency in an online user study, even though the ratings for these were equal to those for more elaborate explanations when it came to unexpected robot actions. 
    These studies highlight the role of transparency and explainability. Indeed, in our study the heuristic-based ToM approach allows us to provide a suggestion that is grounded in the user's moves history but can also be explained in terms of such user's past behaviour. 
    
    Several approaches have sought to simulate ToM through scripted interactions or by having humans control the robot's actions. 
    According to~Romeo~\etal\cite{Romeo_roman22}, the use of ToM affects users' trust in cooperative tasks. They conducted an online user study in which they assessed the impact of three robot behaviours in the context of solving a maze: one neutral, one that explained its reasoning using technical terms and the last one using ToM. Similar results were also found by Mou~\etal\cite{Mou_ROMAN20}. Likewise, Rossi~\etal\cite{Rossi_roman22} provided evidence of how a robot endowed with ToM abilities could be perceived as more competent and therefore more trustworthy even when the latter committed some errors. 
    An interesting work is that by Barchard~\etal\cite{Barchard_THRI20}, in which they combined concepts such as beliefs and intentions with psychological constructs such as behaviours, cognitions, and emotions. Their online study showed that perceptions of a robot's ToM abilities are correlated with their willingness to use the robot.
    Nonetheless, those studies have relied on a human wizard to simulate the robot's ToM abilities, which makes their systems too complex for integration into an autonomous robotic system, particularly in real-world scenarios.
    
    To address the limitations of human-operated simulations, researchers have developed computational models that enable robots to autonomously represent and reason about a user’s internal states and beliefs.
    Nikolaidis~\etal\cite{Nikolaidis_IJRR15} explored how to use a cross-training method (team members can switch their roles to learn shared plans) in the context of human-robot teaming by proposing a computational framework for the robot's mental model. They found that individuals who interacted with a robot that had the capacity to reason about their intention performed better in a collaborative task.
    Shvo~\etal\cite{Shvo_iros22} introduced an algorithm that used epistemic planning-based techniques to enable a Pepper robot to act proactively based on its ability to infer user's intents. With their results, they demonstrated how a robot with ToM abilities is perceived as more useful and socially competent than one without such an ability. 
    Buehler~\etal\cite{Buehler_iros22} investigated when and what information the robot needed to provide the users in a collaborative scenario. They proposed the concept of ToM-based Communication that decided what and when to share information based on its relevance and inferred human beliefs.
    Another cognitive architecture, called Thrive, was proposed by Patacchiola~\etal\cite{Patacchiola_IEEETC22}. 
    Thrive was based on biological insights involving midbrain trial-and-error learning and emphasised the importance of ToM for empathetic trust. The system combined an actor-critic framework to stabilise self-organizing map weights, incorporating Bayesian networks for cost measurement and was applied to the iCub humanoid robot, replicating psychological experiments (sticker-finding task, object-name learning) to align with real data. 
    Bayesian models of Theory of Mind (ToM) have been shown to provide robust frameworks for inferring human beliefs, desires, and intentions. For instance, Baker~\etal\cite{Baker_aaai11}  introduced a Bayesian framework for modelling ToM, enabling systems to attribute both beliefs and desires to human agents based on observed behaviour. The proposed framework uses meta-learning to build ToM models of the agents. 
    The system was tested in classic ToM tasks but not in a real-world setting. Lee~\etal\cite{Lee_hri20}
    presented a computational framework for modelling non-verbal communication during human-robot interactions. Using a Bayesian ToM, they modelled dyadic storytelling interactions, where the storyteller influences and infers the listener's attentiveness through speaker cues, while the listener conveys attentiveness using responses. The framework treats the storyteller's role as a Partially Observable Markov Decision Process (POMDP), while the listener's actions are modelled using a Dynamic Bayesian Network (DBN) with a myopic policy.
    Alanqary~\etal\cite{Alanqary_amcss21} explores how human goal inferences can be modelled by accounting for mistakes in goals, plans, and actions within a BToM framework. They presented an extended model of boundedly rational agents, where mistakes arise due to confusion between similar goals, resource-bounded planning, and execution errors. 
    
    While most of these works coupled the ToM abilities of the robot with its decision-making, others decoupled it separating the ToM layer from the decision-making layer. In the work presented by Devin~\etal\cite{Devin_HRI16}, the robot first constructed a ToM representation of the human's beliefs and goals. Only after this inference was the decision-making module activated, which used the inferred mental state to adapt its behaviour when executing human-robot shared plans. Likewise, Görür~\etal\cite{Gorur_hri17} presented an architecture in which the system first builds a stochastic model of the human's beliefs over possible states. Then, it integrated it into the shared-plan generation.
    Despite their results, these studies have some limitations: i) the robot's ToM capabilities were assessed in very controlled environments e.g., lab settings or online studies, and ii) the studies relied on convenience populations (e.g., students), (iii) the ToM happens at a higher level of abstraction and then used to reshape the robot's actions. In the presented work, we do not aim to build a computational model of the human instead we aim to evaluate our two-layer architecture and how we can decouple actions learning from its implementation. Our heuristic-based ToM operationalises the action of assistance and provides an explanation of why that has been selected. Finally, we evaluate our system in the wild with untrained users with no/limited experience with robotic technologies.

\section{The Memory Game Task}
\label{sec:memory_game}
The cognitive task we used is the classical Memory Game, also known as Concentration which is designed to assess players' memory and concentration skills. Players are typically presented with a set of cards that have images on one side and a uniform back on the other. The cards are placed face down, and players take turns flipping over two cards at a time, trying to find a match. A move is considered as two turns (two cards are flipped). 
The objective of the game is to find all pairs, making as few mistakes as possible in the shortest amount of time.
In line with our previous study~\cite{Andriella_IJSR20}, we decided to define a deck of 24 cards. This number seems to be a reasonable compromise in terms of difficulty in justifying the additional support of the robot in solving the task. 

\begin{figure}[t!]
    \centering
    \includegraphics[width=.48\textwidth]{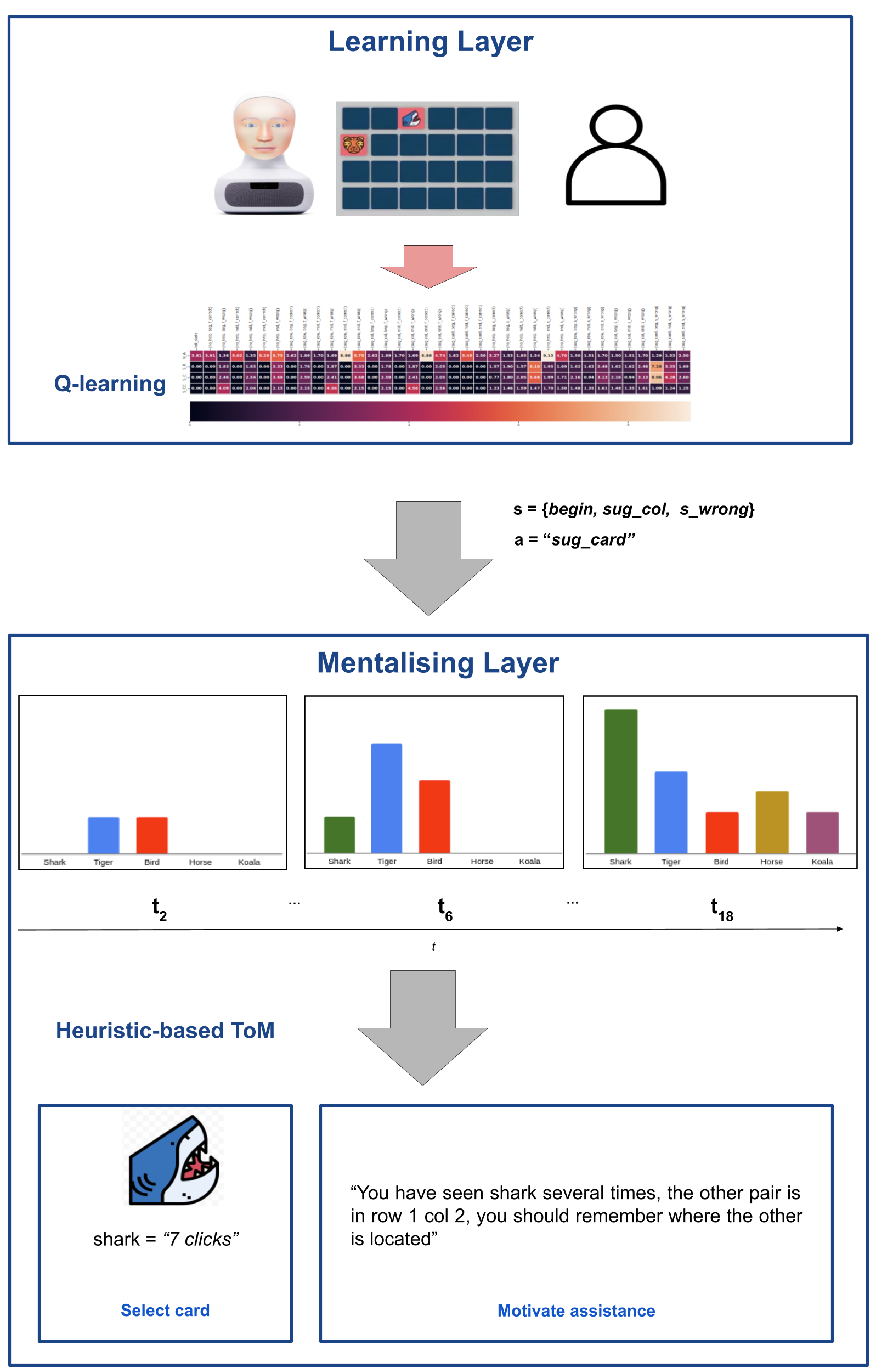} 
    \caption{The two-layer architecture. Firstly, the robot selects the action of assistance that matches the user's performance ($a=sug\_card$) in a given state (${s\_begin, sug\_col, s\_wrong}$). It does so, in the learning layer, by accessing the q-table learnt offline by combining interactions generated in simulation and from previous data.  Next in the mentalising layer, the heuristic-based ToM based on the user's previous moves, estimates the card that might lead to a match (e.g., shark) and then explains why that was selected.}
    \label{fig:architecture}
\end{figure}

\section{Combining Reinforcement Learning with Theory of Mind}
\label{sec:framework}
The main objective of this study is to develop a hierarchical computational approach to generating the socially assistive behaviour that best matches the user's preferences and translating them into meaningful assistance by leveraging the system ToM's abilities. 
Specifically, we defined two layers of abstractions (see Figure~\ref{fig:architecture}). The upper layer employs an RL algorithm to learn the degrees of assistance (see Section~\ref{sec:learning_adaptive_behaviour}) to provide to the user in a given state (see Section~\ref{sec:modelling_user}). The lower layer uses a heuristic-based ToM to infer the user's intended strategy and operationalises the assistance selected by the RL. Additionally, it also provides an explanation of the rationale behind its choice (see Section~\ref{sec:mentalising_user_ability}). It is important to mention that the two layers are completely independent. This has the following advantages: (i) it allows us to learn the robot's assistance at a high level of abstraction, separating it from its implementation; (ii) using a heuristic-based ToM that is not fully integrated into the decision-making system allows us to maintain transparency in the process and provide an explanation for the system's implemented action.

\subsection{Learning to Adapt}
\label{sec:learning_adaptive_behaviour}  
Data-driven approaches require a significant amount of data to learn a reasonable policy. However, collecting this data by exposing users to long interactions with the robot can be difficult and sometimes not feasible. Additionally, during the initial stages, data-driven methods may need to explore undesired states that could potentially hinder the participants and impact their acceptance of the robot. This problem poses a significant obstacle to the use of data-driven approaches in real-world scenarios~\cite{Andriella_roman19}.

This work builds upon our previous work~\cite{Andriella_IJSR20} in which a robot was employed to provide adaptive assistance to users playing a memory game. In~\cite{Andriella_IJSR20}, the robot's assistive behaviour was based on a simple but effective probabilistic model and was trained from data collected by observing humans in the same context. However, the collected data were very limited, and in some states, the robot did not have enough information to make a decision.
In this work, we go a step further, trying to address that limitation and the related cold start problem. We propose a temporal difference (TD) algorithm, Q-learning~\cite{Sutton_MIT98}, to train the robot in finding the optimal policy by combining a data-driven approach, using data from~\cite{Andriella_IJSR20}) and a knowledge-driven approach, building a model of an imperfect player (see Section~\ref{sec:modelling_user}).
We employed Q-learning as it is one of the most used approach from previous studies for providing adaptive robot's behaviour. With respect to more sophisticated RL algorithms like DQN, Proximal Policy Optimization (PPO), or Actor-Critic methods, it is considered better due to its simplicity, robustness, and general applicability, especially in environments with discrete action- and state-space.
Here, the optimal policy is the one that leads the user to complete the game with minimal assistance from the robot.
Differently from previous works, we are not only interested in providing assistance after the first card has been flipped out but also before the player has to flip out any cards. This adds another layer of complexity to the learning algorithm that has to learn socially assistive behaviour in two different states.
    
We formalise the task as a Markov Decision Process (MDP). The action space (\textit{A}) is discrete, and the robot has 4 actions available ($A=\{no\_help, sug\_col, sug\_row, sug\_card\}$). An example of those actions is reported in Table~\ref{tab:assistive_behaviour}. The state space (\textit{S}) consists of the following variables: the game state $GS=\{begin, middle, end\}$ where $0<beg<4$, $4\le mid<8$ and $8\le end<12$ found pairs; the assistance provided by the robot in the previous interaction $RS$=$\{no\_help,$ $sug\_col,$ $sug\_row,$ $sug\_card\}$; and the outcome on the previous turn $US$=$\{f\_correct, f\_wrong, s\_correct, s\_wrong \}$ (where, $f\_$ and $s\_$ stand for first and second card and $correct$ or $wrong$ for its outcome).

Note that in this way, we can distinguish whether the assistance was provided on the first or second card. Therefore, we have $|GS| \times |RS| \times |US|$ = 48 game states.
    
The reward function (\textit{R}) is defined dependably on whether the robot assistance is provided before flipping a card ($r_{1}$) or after having flipped a card ($r_{2}$).
In the case the user flips the first card, we defined $r_{1}=\{\hat{a}/nf\}$ where $\hat{a} \in \{ 10, 0.2, 0.1, 0.025 \}$ and $nf$ is the number of flips before finding a match. Note that the values of $\hat{a}$ were empirically selected in simulation to preserve the user's concentration as per previous work~\cite{Hemminghaus_hri17}. To do so, we decide to provide a higher reward when the robot does not take any action, and only intervene when the number of mistakes can have a significant impact on their performance and positive attitude towards the task. Hence, the lower the assistance and the number of flips is, the higher will be $r_{1}$. 
Similarly, when the user has already flipped the first card, the reward function $r_{2}$ is defined depending on whether the robot provides or not assistance. That is to say that, if $a = no\_help$ then $r_{2}=\frac{\hat{a}}{(nf*\hat{gs})}$, which decreases the reward over time. If $a \in \{sug\_row, sug\_col, sug\_card\}$ then $r_{2}=\hat{a}*(nf*\hat{gs})$, which increases the reward over time ($\hat{gs} \in \{3, 2, 1 \}$).
The rationale behind $r_{2}$ is slightly different from $r_{1}$.
In $r_{2}$, we decrease the reward in case of no assistance to prevent the robot from being stuck with a high-reward action without providing assistance when the player cannot find a match. Overall, the way we define $R$ has to do with our main objective of avoiding boredom and frustration while playing the game. Indeed, if the robot intervenes too often, the user might not feel sufficiently challenged and could lose interest. On the other hand, if the robot does not provide support when needed, the user might experience negative feelings for not being able to find a match. The importance of the \textit{no\_help} action in previous work has often been overlooked \cite{Hemminghaus_hri17}. Intervening constantly during the game might break the user's concentration, which is why it is crucial to provide assistance only when it is needed~\cite{Winkle_rss20}. 

We solve the MDP using Q-Learning.
The goal of the Q-learning is to find the optimal policy $\pi^{*}$ that maximises the cumulative expected discounted reward $R$. The policy is formulated as a tabular matrix called $Q(s, a)$ in which $a$ $\in$ $A$ and $s$ $\in$ $S$.
To select actions over time, we used the $\epsilon$-greedy algorithm. 
The Q-learning requires setting three hyperparameters: the learning rate $\alpha$, the discount factor $\gamma$ and the exploitation-exploration factor $\epsilon$. 
To make the parameters more effective and dynamic over time, we set $\epsilon$=1 / $\alpha$=0.1 for the first episode and decrease / increase them after each episode using a decay / grow rate. In this way, in the beginning, the Q-learning will be free of exploring, updating constantly its q-values, while the time passes, it will become more confident and consider more the learnt policy by exploiting what it has learnt so far ($\epsilon$=0.1, $\alpha$=0.95). On the other hand, the $\gamma$ value was fixed to 0.8. 

\begin{table*}[]
\centering
\caption{Examples of assistive behaviour. 
The \textit{Inferred Information} refers to the knowledge that the robot has about the game. The last column offers an example of how such information is used to assist the player.}
\resizebox{\textwidth}{!}{%
\begin{tabular}{llll}
\hline
\textbf{Action} & \textbf{Card} & \textbf{Inferred information} & \textbf{Implementation Example} \\ \hline
 & $1^{st}$ & Both locations have been visited once & \begin{tabular}[c]{@{}l@{}}``You have seen both locations of \textit{shark} once. Let me refresh your \\ memory: row 1 and col 2.''\end{tabular} \\
\textit{sug\_card} & $1^{st}$ & One location never visited, the other visited multiple times & \begin{tabular}[c]{@{}l@{}}``You have clicked \textit{shark} several times. Click the one in row 1 and \\ col 2, you should then remember where the other one is located.''\end{tabular} \\
 & $1^{st}$ & Both location visited multiple times & \begin{tabular}[c]{@{}l@{}}``You have often seen both locations of \textit{shark}. The one less visited \\ is located in row 1 and col 2. In this way, you should make a match!''\end{tabular} \\
 & $1^{st}$ & One location visited multiple times, the other has never been visited & \begin{tabular}[c]{@{}l@{}}``Are you looking for a particular card? Well, then try row  1 and col\\  2. Surely you remember the other location!''\end{tabular} \\  
\hline \hline
 & $2^{nd}$ & Both locations have been visited once & ``You've seen this card before. I remind you that it is located in row 1.'' \\
\textit{sug\_row} & $2^{nd}$ & One location clicked multiple times, the other has never been visited & ``This card again? You're struggling on this one. Well, then try row 1.'' \\
 & $2^{nd}$ & Both locations visited multiple times & \begin{tabular}[c]{@{}l@{}}``You have seen both locations of \textit{shark} more than once. Try to \\ remember at what location of row 1 the other card is located.''\end{tabular} \\
 & $2^{nd}$ & Current location only visited once, the other has never been visited & \begin{tabular}[c]{@{}l@{}}``You haven't seen a \textit{shark} before, so let me help you: try the third \\ row.''\end{tabular} \\ \hline
\end{tabular}%
}
\label{tab:assistive_behaviour}
\end{table*}

\subsection{Modelling the User}
\label{sec:modelling_user}
Modelling the user is fundamental for designing robot behaviours that match their needs and preferences~\cite{DiNapoli2023}. 
In this work, we model the user taking into account: (i) the task, by defining users with imperfect memory based on the findings of Vellemant~\etal\cite{Velleman_AMM13} and (ii) the data gathered in our previous work~\cite{Andriella_IJSR20}. In~\cite{Velleman_AMM13}, the authors provided some interesting formulas to estimate the optimal performance of a player with perfect memory. They say that the expected number of moves is $\approx$ 1.61 $\cdot$ $n$ where $n$ is the number of expected matches. In our case, this number is around 19.
Furthermore, the expected number of flips, before two matching cards are identified, is $\sqrt{\left(\pi \cdot n\right)}$. In our case, this number is around 8 moves. This information provides us with an upper bound whereby we can model more realistic participants.

Firstly, we design the perfect player according to~\cite{Velleman_AMM13}. Next, we model a player with imperfect memory by hypothesising that the probability of the user making a match $P(U)$ depends on the number of cards flipped so far, $NF$ and the number of matches $NM$. In this way, the more the user plays ($NF$ and $NM$), the higher will be their chance to find a pair.  Using linear regression, we analyse data from our previous study~\cite{Andriella_IJSR20} to estimate how player performance changes over time as a function of these variables. In the following paragraph, we provide some details on the implementation. 

When a user flips their first card, they randomly select it with the goal of exploring new areas of the board while also exploiting what they have already seen. To do this, we use a simple heuristic that balances between moves exploring the board and selecting previously seen cards.
In this latter case, the probability of finding a match is increased because the simulated player has already seen some cards. We calculate this probability as $P(seen)=(1-d)^{NF}$, where $d$ is the decay rate. In this way, we simulate the user's memory of a card. The higher the value of $d$, the more forgetful the user becomes.         
If the user flips a second card, the probability of making a match is:
\begin{equation}
\label{equ:reward}
\begin{aligned}
    P(U) = \begin{cases}
    \text{if not seen} & \frac{1}{(NP-NM)}
    \\ \text{if already seen} & \frac{P(seen)*NF}{(NP-NM)}
    \end{cases}
\end{aligned}
\end{equation}
, where $NP$ is the number of pairs.
    
It is important to note that the probability $P(U)$ is also affected by the assistance provided by the robot. That is to say, if the user is provided with $a$=$sug\_row$ or $sug\_col$, then the probability of selecting the correct card is increased proportionally according to the number of cards in the row/column (e.g., 1/4 for col and 1/6 for row). Finally, if the user is provided with the card ($a$=$sug\_card$), their probability will be, obviously $P(U)=1$.

\subsection{Mentalising Users Ability}
\label{sec:mentalising_user_ability}
    
The mentalising layer is responsible for making inferences on the actions $A$ provided by the robot when the user has not yet flipped a card. 
The game can be approached using three specific moves: \textit{0-unknown}, \textit{1-unknown}, and \textit{2-unknown}. 
A \textit{0-unknown} move entails flipping over known cards that the player assumes to be a pair. In the event the assumption is incorrect, no new information is gained.
In a \textit{1-unknown} move, one unknown card is flipped over and matched with a known card. If a match is established, new information is gained. Otherwise, a known card is flipped over that does not match.
Lastly, a \textit{2-unknown} move involves flipping over two unknown cards to match with known cards. If a match is established, new information is gained. If not, the player must repeat the move.
    
A possible assistance strategy from a user would be in the case of 1-unknown. Indeed, it is helpful to keep in mind that if they already know the position of one card, their chances of success are increased.  However, if the user cannot remember that they have already seen that card in a different location, they might end up flipping another unknown card (2-unknown). 
Therefore, the system keeps track of the cards flipped so far by the user, sorting them according to the number of flips and in which turn. Then, when the $Q(s, a)$ matrix suggests an assistive action such as $sug\_row$, $sug\_col$, or $sug\_card$ on the first card, the mentalising layer selects the location of one of the most clicked cards in the recent history. It then suggests the corresponding matching card, which is unknown to the user. The rationale behind this approach is as follows: if the user has already seen a card multiple times, they are likely looking for a specific match. Therefore, the system suggests the location hint ($sug\_row$ or $sug\_col$) or the card ($sug\_card$) of the matching card that the user has not seen yet or has seen less frequently. Additionally, the robot not only provides this information, but it also offers a simple explanation for why it has made the suggestion, demonstrating its mentalising abilities (see examples in Column ``$1^{st}$'' of Table~\ref{tab:assistive_behaviour}).
    
Once the user has flipped out the first card, depending on the latter, the system provides assistance according to the $Q(s,a)$ (see Column ``$2^{nd}$'' of Table~\ref{tab:assistive_behaviour}). 
It is worthwhile noticing that the assistance given for the second card depends on both the card recently flipped by the player and its history. Indeed, the robot always provides assistance in a way that could potentially lead to a match with the already flipped card, regardless of whether they have seen it before or not.

\begin{figure}[t!]
        \centering    	
        \includegraphics[width=\linewidth]{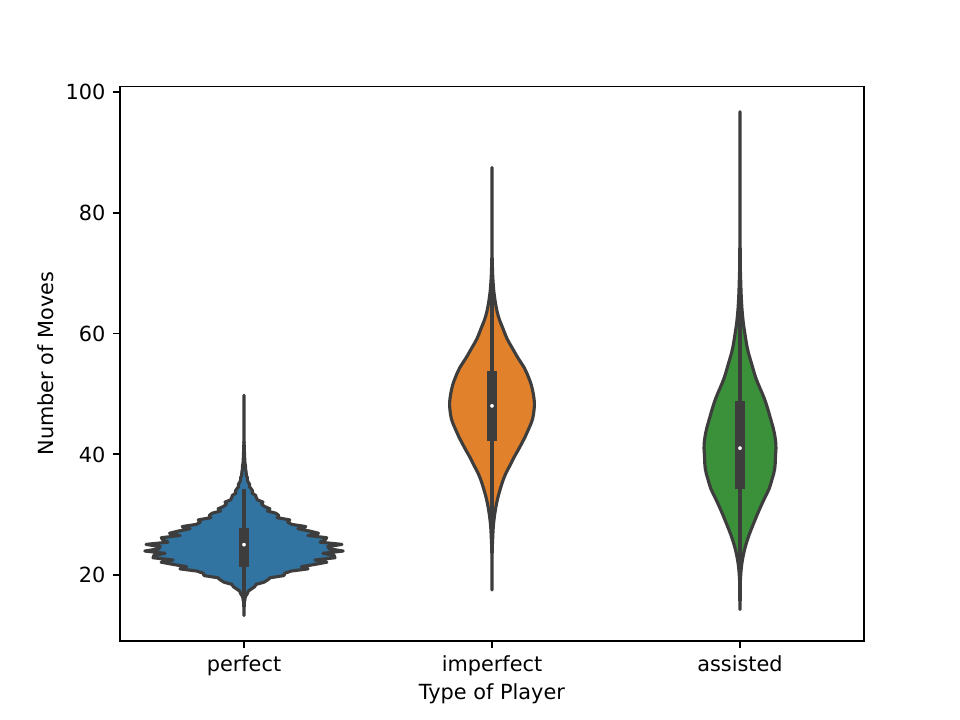}
        \caption{The figure shows the results in simulation of a perfect player ($M=25.1$ $SD=3.7$), an imperfect player ($M=48.15$, $SD=7.55$), and an imperfect player playing assisted by the Q-learning agent ($M=41.73$, $SD=8.83$), respectively.} 
        \label{fig:simulation}
\end{figure} 

\begin{figure*}[t!]
        \centering    	
        \includegraphics[width=\linewidth]{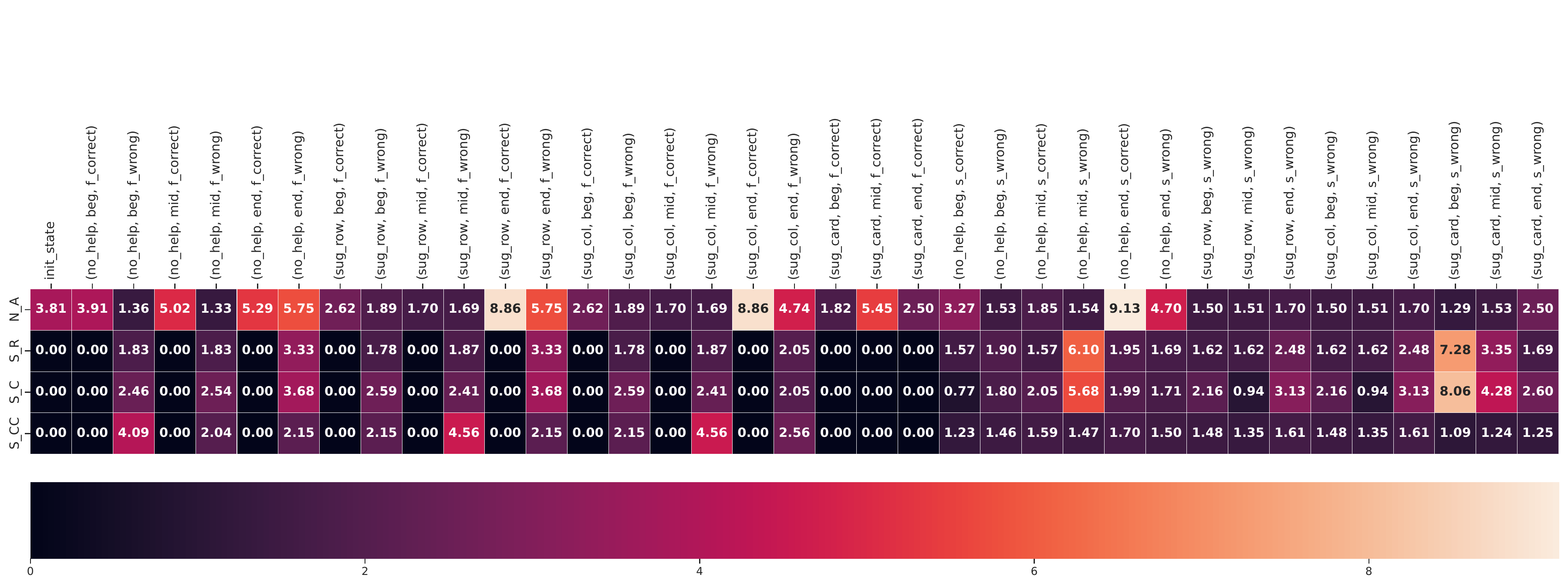}
        \caption{The figure shows the Q-matrix learnt from the RL agent. On the y-axis, the four actions: note that $N\_A$ stands for $no\_action$, $S\_R$ stands for $sug\_row$, $S\_C$ stands for $sug\_col$, $S\_CC$ stands for $sug\_card$. On the x-axis, the states. Note that we did not include those states that the agent never visited.}    	
        \label{fig:learnt_policy}
    \end{figure*} 

 \subsection{Results from Simulation}
Here, we summarise the findings obtained in the simulation as a result of modelling the perfect player, the imperfect player, and the imperfect player assisted by the Q-learning agent, respectively.
The results are shown in Figure~\ref{fig:simulation}. The perfect and imperfect players differ for about 23 moves, while the assisted imperfect player performs slightly better thanks to the hints offered by the robot. Results are aligned with our previous findings~\cite{Andriella_IJSR20}. Note that we did not model the complexity of the game (similarity of the cards), which in a real scenario can have an impact on the players' performance. Additionally, we did not consider the user's trustworthiness regarding the robot's assistance. Therefore, the user accepted any assistance provided by the robot.

The resulting policy learnt from the system in the case of the imperfect player is reported in Figure~\ref{fig:learnt_policy}.
Note that to make the matrix more readable, we removed the states the RL agent never visited, e.g., s($sug\_card$, $beg$, $f\_wrong$). 
It can be observed how the system is able to distinguish between the two states in which it is requested to assist.
On the first flip, the system tries not to intervene if the user performs well and increases its degree of assistance when the player starts making only a few mistakes. The assistance is increased at the beginning and decreases over time as soon as the player starts finding matches.
On the second flip, the system reduces the time provided on the first card for exploring the board and provides the user with assistance more often.

\section{Experimental Design}
\label{sec:methodology}
The study was set up as a between-subject study, in which we manipulated the robot's ability to mentalise the user's intent during the game. As in our previous works~\cite{Andriella_UMUAI22, Andriella_IJSR20}, we envisaged the robot as an assistant that aids the users during the memory game. Ethical approval was obtained from the ethical committee of the University of Naples Federico II.
Each participant was randomly assigned to one of the two conditions: 
\begin{itemize}
    \item \textit{without\_mentalising abilities} (\textbf{noToM}):  On the first card, the robot did not make any reasoning on the cards previously explored by the player. Once an assistive action was selected by the Q-learning, the robot instantiated it randomly without providing any motivation (\textit{``try to flip a card in row 2 col 3"}).
    For the second card, assistance was provided without any explanation. For example, it correctly identified the row ($sug\_row$) based on the first card (\textit{``The matching card is located in row 2"}).
    \item \textit{with\_mentalising abilities} (\textbf{ToM}): On the first card, the robot provided assistance according to the mechanism described in Section~\ref{sec:mentalising_user_ability}. 
    For instance, if the action was $sug\_card$, the robot would select the card based on the user's previous history inferring their next move.
    Therefore, it might suggest the other location of ``shark" providing a motivation of why it has suggested it (\textit{``You have seen shark several times, you should remember the other location"}). For the second card, assistance was provided with an explanation. For example, it correctly identified the row ($sug\_row$) based on the first card (\textit{``You’ve seen this card before. I remind you that it is located in row 2"}). 
    \end{itemize} 

It is worthwhile mentioning, that in both conditions, the robot was endowed with the same Q-learning algorithm (see Section~\ref{sec:learning_adaptive_behaviour}). 
The only difference was how the assistance was provided (without or with ToM).

To demonstrate the presence or absence of an effect, we analysed the data using a T-test (normality checked with Shapiro-Wilk test) and Mann-Whitney U test. For N=56, we estimated an effect size of $d=0.35$ with $0.84$ power at an $\alpha$ level of $0.05$. 
We recognise that conducting multiple comparisons without applying a correction to the alpha value increases the likelihood of Type I errors (false positives). Given that this study is exploratory in nature and seeks to provide initial insights into the impact of ToM on HRI, we opted not to apply the Bonferroni correction in this analysis, as it would risk being overly conservative and potentially obscure meaningful trends.
To foster reproducibility, we have open-sourced our code\footnote{\url{https://github.com/Prisca-Lab/Q-learning_Concentration}}.

    \subsection{Hypotheses}
    We formulated the following research hypotheses:
    
    \begin{enumerate}
    [start=1,label={\bfseries H\arabic*:}, leftmargin=*,align=left]
        
        \item participants assisted by a robot endowed with ToM ability solve the game with fewer mistakes and in a shorter time compared to those who are assisted by a robot without such ability.


        \item Participants assisted by a robot endowed with ToM ability follow its suggestions more often compared to those who are assisted by a robot without such ability.

        \item Participants assisted by a robot endowed with ToM ability capabilities believe the robot is better at predicting their beliefs and intentions compared to those who are assisted by a robot without ToM capabilities.
    \end{enumerate}
    
   In H1, our goal was to evaluate whether the robot equipped with ToM had any effect on the participants' performance. Similarly, in H2, we aimed to assess the quality of assistance provided by the robot capable of ToM. Lastly, in H3, we investigated the users' perception of the robot.

\subsection{Experimental Setting}
The experiment was carried out during a technology fair, that involved mainly young students from high schools and their relatives. On a desk, we set up a computer from which participants played the game, and on a side with respect to the computer we located the robot (see Figure~\ref{fig:user_playing}). As a robotic platform, we employed the Furhat robot\footnote{\url{https://furhatrobotics.com/}}. The experimenter was seated at the same table but out of the field of view of the participants to not interfere with the experiment and distract them. 

\subsection{Apparatus}
The experimental apparatus comprised a Furhat robot and a computer-based interface, both integral components of the interactive game employed in this study. The robot, developed by Furhat Robotics, played a pivotal role as the interactive agent in the experiment. This anthropomorphic robot, standing at approximately 60 centimetres in height, is designed to simulate natural face-to-face interactions with humans. It is endowed with a high-resolution facial display, capable of expressing a wide array of emotions and non-verbal cues. The robot's non-verbal social cues were constrained to the standard ones provided by the SDK. The robot could display, happy, confused or sad faces depending on whether the user made a correct or a wrong move.
The memory game\footnote{\url{https://github.com/yunkii/animal-memory-game}} was played on a laptop computer of 15 inches. Participants interacted with the game by means of a mouse.

\subsection{Procedure}
\label{sec:procedure}
The experiment was conducted in a booth where hundreds of visitors were wandering around. Upon arrival,  the experimenter introduced the experiment, providing a brief overview of the study's purpose and procedures.

If the participant expressed interest and agreed to participate, they were asked to provide informed consent by filling in a consent form. The consent form included information about the study's objectives, the experimental tasks involved, the potential risks and benefits, and the assurance of confidentiality. The participants were given ample time to read and understand the contents of the consent form before signing it.

After obtaining informed consent, the participants were instructed to proceed with the experimental task. They were given a clear explanation of the game they would be playing and were provided with any necessary clarifications to ensure their understanding. 
Next, they were given up to 5 minutes to familiarise themselves with the game and with the robot. In this stage, the game was a different one with respect to what they played in the experiment and the experimenter would address any question raised by the participant.
After that, the player could play the memory game. The experimenter explained that the objective of the game was to minimise the number of mistakes and complete it in the shortest time possible with the help of the robot. 
The estimated duration for completing the game was approximately 5 minutes. 

After finishing the game, participants were asked to fill in a set of questionnaires. These questionnaires aimed to assess their subjective experience with the robot encountered during the game. 
The estimated time required to complete the questionnaires was approximately 8 minutes.

\subsection{Participants}
A total of sixty participants were initially included in the experiment. However, after a careful examination of the data, four participants were identified as outliers due to their performance deviating by more than 2 standard deviations (2$\sigma$) from the rest of the participants. As a result, these four participants were excluded from the analysis, leaving a final sample size of fifty-six participants. From those 28 were assigned to the noToM group and the remaining 28 to the ToM group.

Among the remaining participants, nineteen identified themselves as female, while thirty-seven identified as male. The participants' ages ranged from 18 to 54 years, with a mean age (M) of 25.98 and a standard deviation (SD) of 8.76.

Regarding prior experience with robots, fourteen participants reported having no previous experience with robots. Twenty-two participants acquired knowledge about robots through exposure to films, books, or series. Nine participants had prior experience with robots at fairs. Six participants owned a robot such as Roomba for domestic use. Additionally, five participants reported having professional involvement with some form of robotic technology.

The recruitment process aimed to achieve diversity in terms of gender and prior experience with robots, allowing for a more comprehensive analysis of the participants' responses and perceptions.

\subsection{Evaluation Measures}
To address our initial hypotheses, we gathered subjective and objective measures.

Regarding the subjective measures, we collected participants' perceptions of the robot intelligence by administering the Perceived Social Intelligence (PSI) questionnaire~\cite{Barchard_THRI20}. As we were not interested in all the dimensions, we only focused on: Social Competence (SOC), Adapts to Human Cognitions (AC), Predicts Human Cognitions (PC) and Recognizes Human Cognitions (RC). Those measures will help us to evaluate H3.

Concerning the objective measures, we collected participants' performance in terms of number of moves, completion time, average time for a match, and the total number of received suggestions (to address H1). Finally, we also measured the number of suggestions followed and not followed (to address H2). 

\begin{figure*}[t!]
    \centering
    \begin{subfigure}{0.24\textwidth}
        \centering
        \includegraphics[width=\linewidth]{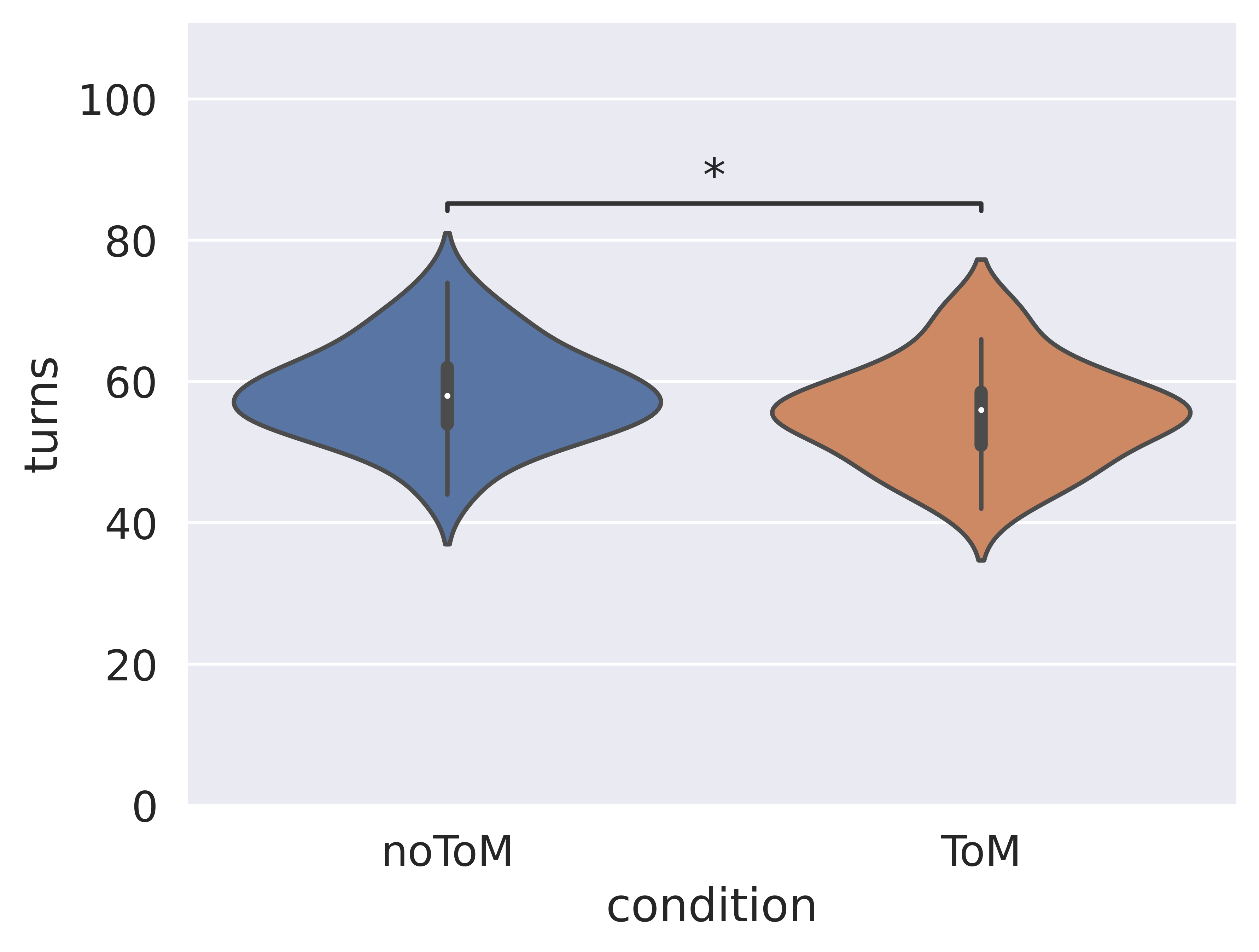}
        \caption{}
        \label{fig:turns}
    \end{subfigure}
    \hfill
    \begin{subfigure}{0.25\textwidth}
        \centering
        \includegraphics[width=\linewidth]{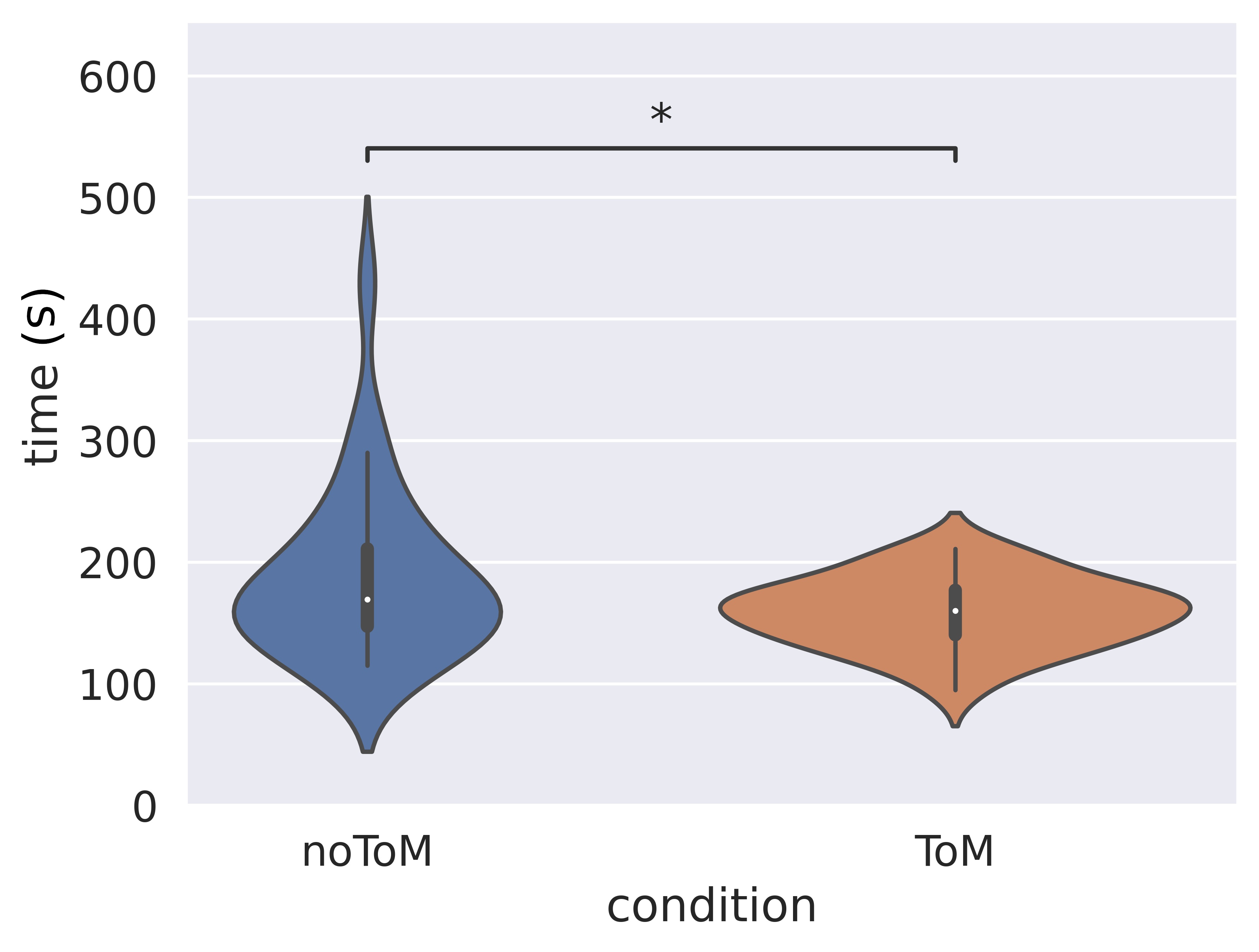}
        \caption{}
        \label{fig:completion_time}
    \end{subfigure}
    \hfill
    \begin{subfigure}{0.24\textwidth}
        \centering
        \includegraphics[width=\linewidth]{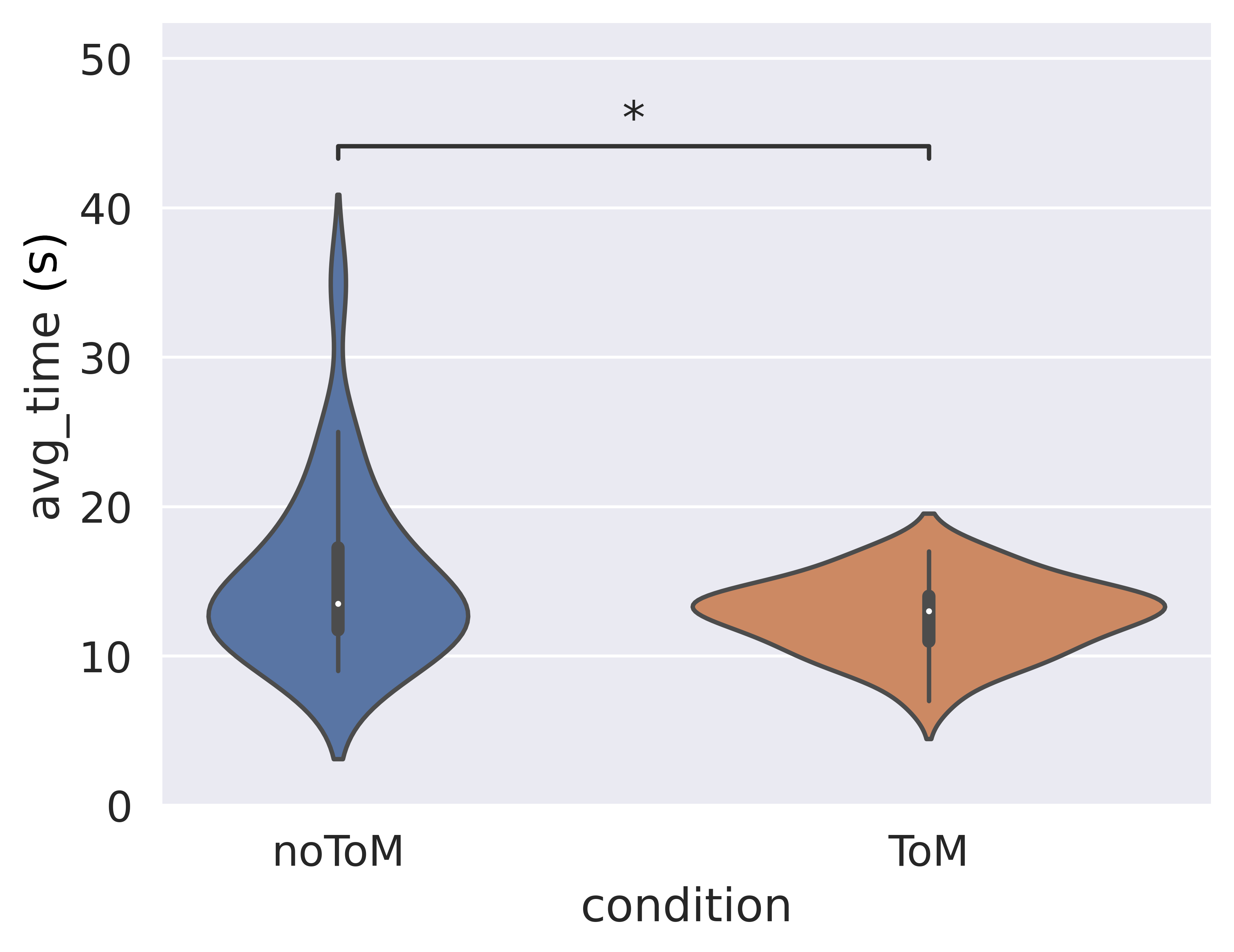}
        \caption{}
        \label{fig:average_time_for_match}
    \end{subfigure}
    \begin{subfigure}{0.24\textwidth}
        \centering
        \includegraphics[width=\linewidth]{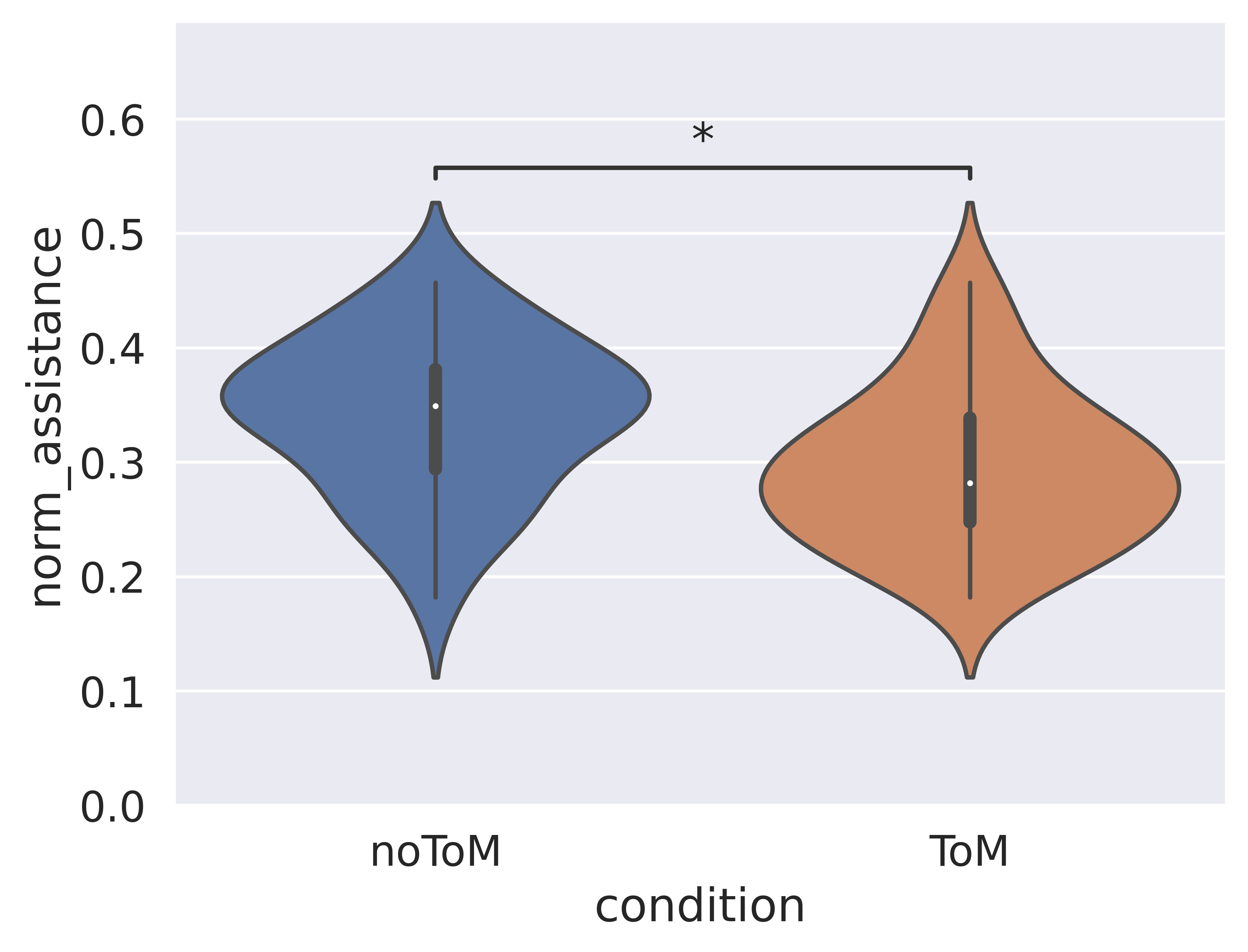}
        \caption{}
        \label{fig:received_assistance}
    \end{subfigure}
    
    \caption{The figure shows (a) the number of turns, (b) completion time, (c) time for a match, (d) the number of assistance received for the participants who belonged to the \textbf{noToM} group (left, blue) and the \textbf{ToM} (right, orange), respectively ($*$ denotes .01 $<$ p $<$ .05).}
    \label{fig:meaningful_assistance}
\end{figure*}

\subsection{Results}
\label{sec:results}
To test hypothesis H1, we run a t-test (data normality assumption was assessed with Shapiro-Wilk-test) with the mentalising ability of the robot as the independent variable and with the number of turns, completion time, the time before a match, and the number of suggestions provided by the robot as dependent variables (see Figure~\ref{fig:meaningful_assistance}). 
Concerning the number of turns, we found that there was a significant effect as participants in the ToM condition ($M$=55, $SD$=7) took fewer moves to complete the game ($p<0.05$, $t$(27)=1.71) in comparison to those in the noToM condition ($M$=58, $SD$=7) (see Figure~\ref{fig:turns}).
Similarly, results indicated that participants in the ToM condition could solve the game in a shorter amount of time ($M$=159s, $SD$=29s) compared to those in the noToM condition ($M$=187s, $SD$=69s, with $p<0.05$, $t$(27)=-2.01) (see Figure~\ref{fig:completion_time}). 
Finally, results showed that participants who belong to the ToM group took less time to find a match ($M$=13s, $SD$=2s) in comparison to those in the noToM condition ($M$=15s, $SD$=6s, with $p<0.05$, $t$(27)=1.9) (see Figure~\ref{fig:average_time_for_match}, where completion\_time/ n\_matches). 
To further understand those results, we also evaluated how much assistance the robot provided in the two conditions. 
For a better comparison, for each participant, we normalised the assistance received with respect to the number of turns (overall\_assistance/n\_turns), summing up the levels of assistance received (overall\_assistance) as follows: N\_A (no assistance) = 0,  S\_C (suggest column) = 0.5, S\_R (suggest row) = 1 and S\_CC (suggest card) = 2. Results showed that participants who interacted with the robot with ToM ($M$=0.27, $SD$=0.07) capabilities were provided with less assistance ($p<0.05$, $t$=-2.4)  than those who played with the robot in the noToM condition ($M$=0.34, $SD$=0.07) (see Figure~\ref{fig:received_assistance}). 

\begin{figure*}[h!]
    \centering
    \begin{subfigure}{0.4\textwidth}
        \centering
        \includegraphics[width=\linewidth]{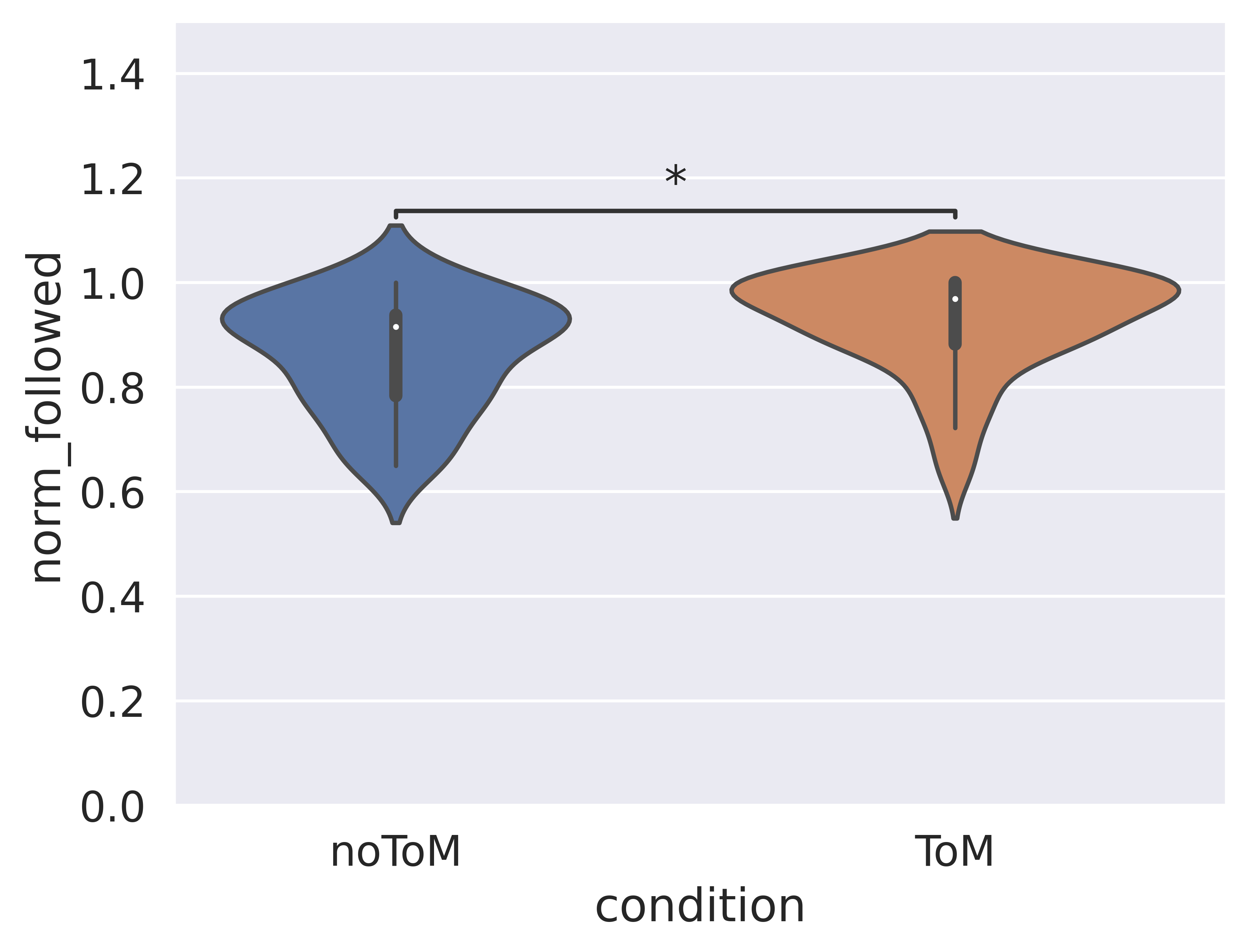}
        \caption{}
        \label{fig:followed_suggestions}
    \end{subfigure}
    \hfill
    \begin{subfigure}{0.4\textwidth}
        \centering
        \includegraphics[width=\linewidth]{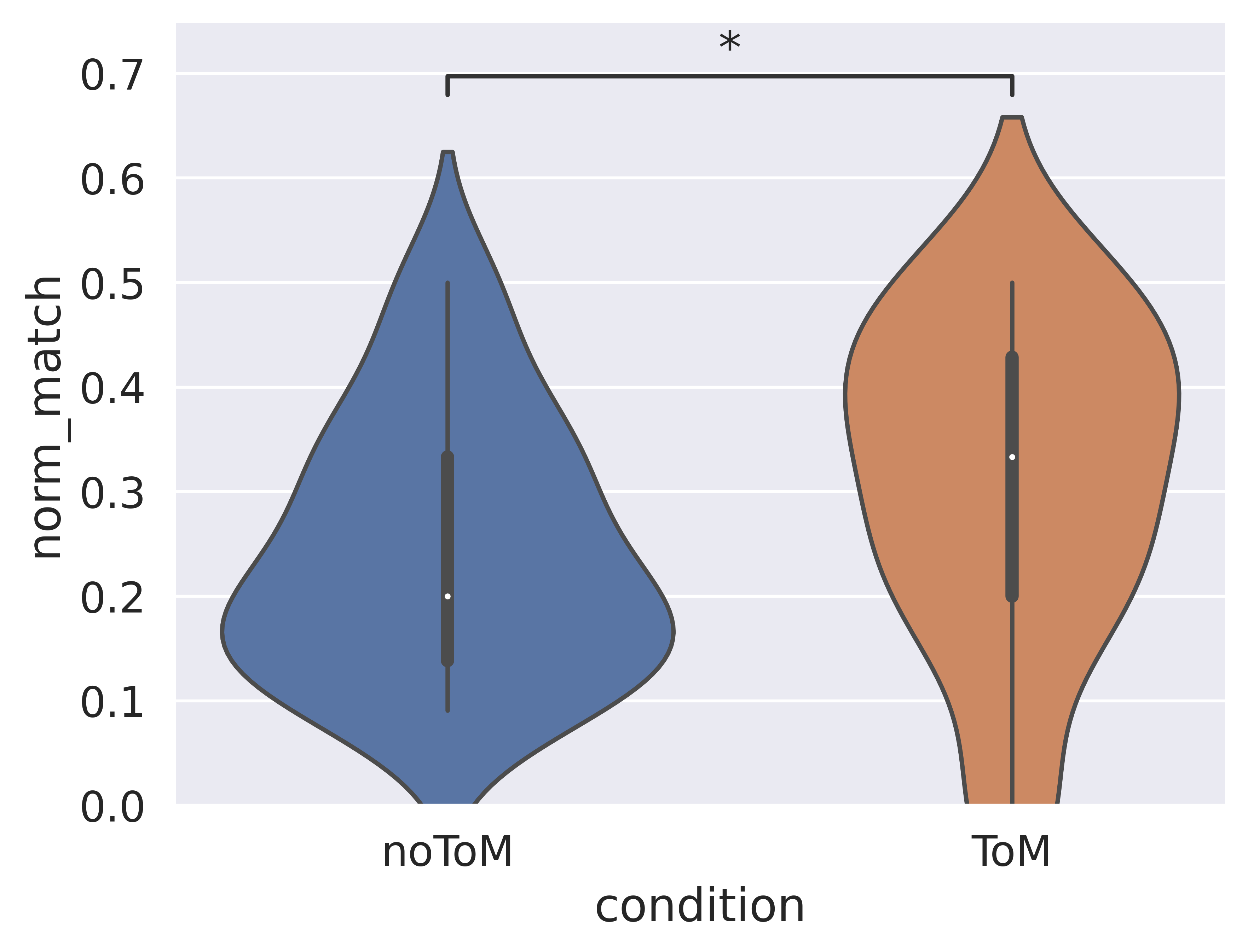}
        \caption{}
        \label{fig:n_led_match}
    \end{subfigure}
    \caption{The figure shows (a) the number of suggestions followed, (b) the number of suggestions followed that led to a match for the participants who belonged to the \textbf{noToM} group (left, blue) and the \textbf{ToM} (right, orange), respectively (n.s. denotes p $>$ .05, $*$ denotes .01 $<$ p $<$ .05).}
    \label{fig:mental_model}
\end{figure*}

To test hypothesis H2, we run a Mann-Whitney U-test with the mentalising ability of the robot as the independent variable and with the number of suggestions followed on the first card and the number of suggestions followed that led to a match as dependent variables (see Figure~\ref{fig:mental_model}). As before, for each participant, we normalised the suggestions with respect to those who were provided during the whole game (n\_followed\_suggestions / total\_suggestions). 
Results showed that participants who played the memory game in the ToM group ($M$=0.92, $SD$=0.09), followed more often the assistance provided by the robot ($p<0.05$, $z$=2.54), in comparison to those in the noToM group ($M$=0.86, $SD$=0.1) (see Figure~\ref{fig:followed_suggestions}). 
Likewise, we normalised the number of cards that led to a match (n\_cards\_match / total\_suggestions). We found that participants in the ToM condition ($M$=0.30, $SD$=0.15) selected more often cards that led to a match ($p<0.05$, $z$=1.94), in comparison to those in the noToM condition ($M$=0.24, $SD$=0.12) (see Figure~\ref{fig:n_led_match}).

\begin{figure}[t!]
    \centering
    \includegraphics[width=\linewidth]{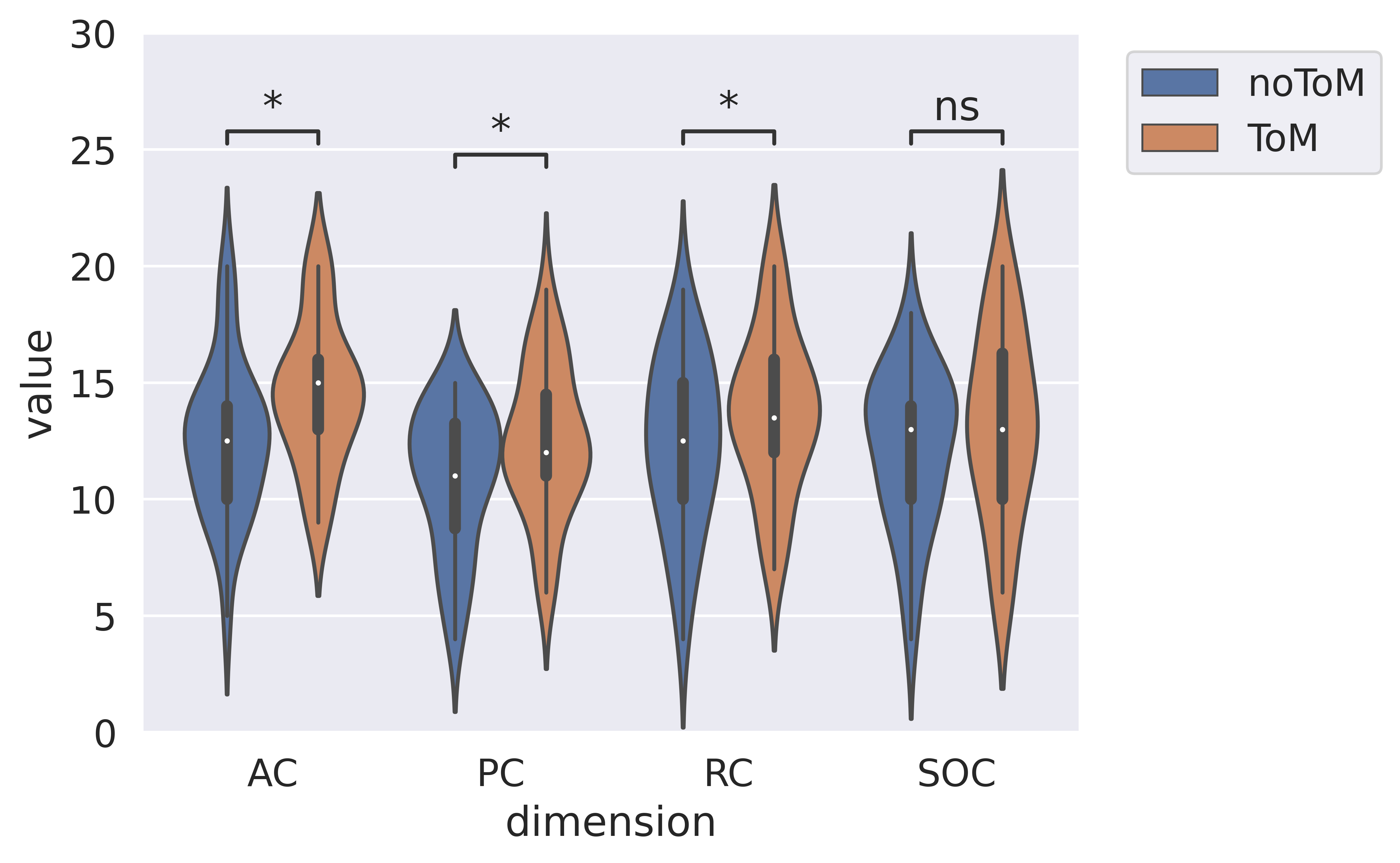}
     \caption{The figure shows the results of Adapts to Human Cognitions (AC), Predicts Human Cognitions (PC), Recognizes Human Cognitions (RC), and Social Competence (SOC) for the participants who belonged to the \textbf{noToM} group (left, blue) and the \textbf{ToM} group (right, orange), respectively (n.s. denotes p $>$ .05, $*$ denotes .01 $<$ p $<$ .05)}
    \label{fig:psi}
\end{figure}

To address our last research hypothesis H3, we analyse the results of the questionnaire  (see Figure~\ref{fig:psi}).
Results from the AC dimension indicated that participants in the ToM group ($M$=15, $SD$=3) perceived the robot as more capable of adapting to their beliefs ($p<0.05$, $t$(27)=-2.10) compared to those in the noToM group ($M$=13, $SD$=3).   
Similarly, results from the PC dimension showed that participants in the ToM group ($M$=12, $SD$=3) perceived the robot as more capable of predicting their actions ($p<0.05$, $t$(27)=1.92) in comparison to those in the noToM group ($M$=11, $SD$=3). Likewise, for the RC dimension, we found that participants in the ToM condition ($M$=14, $SD$=3) perceived the robot with more capabilities of recognising their actions ($p<0.05$, $t$(27)=1.70
) compared to those in the noToM condition ($M$=12, $SD$=4).
Finally, with respect to the SOC dimension, we did not find any statistical significance. Participants who interacted with the robot endowed with ToM capabilities ($M$=13.21, $SD$=4.02) did not score significantly higher ($p=0.17$, $t$(27)=0.93) than those who interacted with the noToM robot ($M$=12.14, $SD$=3.33). 
 
\section{Discussion}
\label{sec:discussion}
In this section, we engage in a comprehensive discussion of our results, focusing on their alignment with our initial research hypotheses and their broader implications. We organised our discussion around the three hypotheses explored in our research.

    \subsection{Hypothesis H1: Impact on Game Performance}
    Our initial hypothesis was that participants who interacted with a robot possessing ToM capabilities would demonstrate better performance than those interacting with a robot lacking ToM. Our findings support this hypothesis across various performance metrics, including number of turns, completion time, and time to find a match.

    Further analysis revealed that the improved performance of participants in the ToM group was not due to receiving more assistance from the robot. The robot provided less assistance to the ToM group. This can be attributed to the offline policy, which was trained to provide more assistance after more failures – a common occurrence among participants in the noToM group.
    Therefore, we can speculate that the quality of assistance received on the first card, rather than the quantity of assistance provided, played a role in the superior performance of the ToM group. These findings seem to be aligned with most of the recent work on robot explainability and its impact on users' understanding by providing more meaningful and context-based information~\cite{Ambsdorf_roman22}. For instance, Matarese~\etal\cite{Matarese_FRAI22} found that when the robot provided suggestions by considering the participants' perspective, they ended up performing better than when they were provided with a less informative suggestion that did not consider the user's point of view.
    Likewise, Devin~\etal\cite{Devin_HRI16} discovered that, when participants were provided with the correct amount of information, they performed better in a shared task.
    Our work is a bit different in that we are not deciding whether to provide information or not, but rather providing information based on the previous history of the participants, making assistance more informative and therefore more helpful for them.  
    
    \subsection{Hypothesis H2: Acceptance and Quality of the Robot Assistance}
    Our second hypothesis centred on the influence of the robot's ToM on participants' acceptance of the assistance, along with their impact on guessing the correct card.
    The results suggest that those participants who interacted with the robot capable of mentalising their strategy tended to follow their assistance more frequently, which, in turn, was associated with a higher likelihood of achieving a match. 
    These findings are consistent with previous work in which robots endowed with some degrees of ToM were deemed more trustworthy to the extent that they could impact people's decision-making~\cite{Rossi_roman22, Romeo_roman22, Mou_ROMAN20}.
    While trust was not explicitly measured in this study, the way the robot provided support in the ToM condition could suggest a potential link to perceived trustworthiness. Indeed, by mentalising the potential strategies of each of the participants, the latter could acknowledge its effectiveness by making matches.

    \subsection{Hypothesis H3: Perception of Robot Competence}
    Our third hypothesis delved into participants' perceptions of the robot's competence, specifically focusing on Adaptation to Human Cognitions (AC), Prediction of Human Cognitions (PC), Recognition of Human Cognitions (RC), and Social Competence (SOC). The outcomes provide interesting insights into the interplay between robots equipped with ToM and participants' perceptions.

    The participant's perception of the robot's social intelligence abilities further supports the more objective findings of H1 and H2. We found significance in the AC, PC and RC dimensions. If we closely examine the statements of these dimensions, we could notice that they assess the robot's adaptive capabilities to the user's beliefs and intents (AC), with its ability to predict the participant's beliefs and thoughts (PC) along with the robot's ability to detect beliefs and thoughts (RC). This is precisely what we aimed for by combining RL with the mentalising layer - designing an adaptive system that, by leveraging its ToM capabilities, can further increase the users' perceived social intelligence of the robot. Our results are aligned with previous research that highlighted how these dimensions are critical factors that could enhance user acceptance of the robot~\cite{Barchard_THRI20}.
    Indeed, Söderlund~\etal\cite{Soderlund_JRCS22} found in their study, that when participants perceived that the robot has ToM capabilities, they assign it a higher degree of usefulness. Likewise, Shvo~\etal\cite{Shvo_iros22} could prove the impact of the robot's ToM on people's perceived utility.

\section{Limitations and Outlook}
\label{sec:limitations}
While our study yielded promising results, it is essential to acknowledge certain limitations that could guide future research. These limitations can be broadly categorized into developmental and methodological aspects.

From a developmental standpoint, some technical constraints affected the current design. One limitation was the offline nature of the policy learning algorithm. Although this ensured that the robot's policy was reliable for gameplay and allowed us to isolate the effects of ToM, the system lacked real-time learning. Future research could explore fine-tuning the robot's behaviour dynamically during interaction to better accommodate individual preferences.
Concerning the current architecture, future work could focus on comparing it with more sophisticated ones that incorporate user mental states such as beliefs, intentions, and strategies within the MDP.
Additionally, the robot's communicative abilities were limited to basic verbal and non-verbal cues. Although this design ensured more control over the manipulated variables and platform independence, it restricted the robot's expressiveness. For instance, a robot capable of pointing could help direct participants' attention more effectively~\cite{Andriella_UMUAI22, Andriella_IJSR20}. Similarly, the robot's spoken responses were predefined and adapted only to specific conditions. This restricted flexibility could be improved by incorporating Large Language Models (LLMs) in future iterations, making the assistance more natural and human-like.
Concerning the timing of the robot's assistance, it was fixed, and we did not explore when the most effective moments for intervention might be. If we use ToM to enable the robot to provide more proactive, context-aware assistance, it could improve the overall user experience~\cite{Cucciniello_HRI23}. Finally, the robot's current set of assistive behaviours was limited to four. Future developments could involve expanding the variety of assistive behaviours, allowing the system to learn more diverse and nuanced responses.

On the methodological front, some factors influenced the scope and generalisability of the findings. 
The sample size, although typical for real-world studies, was relatively limited. 
Moreover, the statistical analyses used in the study might have increased the potential for Type I errors due to the lack of corrections for multiple comparisons. Replicating the results with a larger sample size and employing multivariate analyses could provide more robust insights. The experimental setup posed another challenge. Participants completed the tasks in a national fair setting, where external factors such as noise and distractions may have impacted their concentration and performance. Although all participants experienced the same conditions, the setting could still have affected the outcomes. In addition, this study was an initial exploration of the technology, intentionally evaluated with a broad, non-specific population. In the future, we intend to assess the system with more targeted groups, such as healthy older adults and individuals with cognitive decline, to validate whether the findings hold across different user demographics.
Concerning the experimental manipulation, some potential limitations can be noted. The differentiation between the ToM and noToM conditions in this study could be seen as confounded by the difference in the level of explanation provided by the robot. It is possible that the experimental manipulation was perceived as ``More Explanation vs. Less Explanation" or ``More Transparency vs. Less Transparency" rather than purely a manipulation of ToM. Indeed, previous studies have shown that providing system explanations can positively affect users' perceptions of systems, leading to increased trust, satisfaction, and understanding~\cite{Lim_sigchi09, Vanderwaa_AI21, Conati_AI21}. However, the focus of our study was not on making the system more explainable and transparent but rather on designing an architecture that allowed us to decouple the learning of the assistive action from their implementation. The way we developed the ToM layer enabled the robot to operationalise its action and as a consequence to explain the reason for its decision.
This specific aspect of mentalising is distinct from merely providing more explanations or making the system more transparent. Therefore, although those two constructs may be partial factors in our study, the primary manipulation involves the robot’s ability to understand and adapt to users’ inferred strategies, as evidenced by the participants' higher scores in AC, PC, and RC in H3. These results suggest that the manipulation was indeed focused on ToM.
Finally, regarding the results obtained addressing H2, while the study revealed a potential link between users accepting recommendations from the robot with ToM capabilities, we did not explicitly measure trust. Future studies should investigate whether the robot's adaptive, ToM-based assistance fosters trust, and if so, how it does so.

\section{Conclusions}
\label{sec:conclusions}
In this paper, our primary objective was to investigate how incorporating ToM capabilities into a socially assistive robot can enhance user's performance and perception in a memory game scenario. Our approach hinged on a hierarchical computational architecture consisting of two separate layers. The first layer, namely the learning layer, rooted in RL, involved pre-training a q-learning agent offline to learn helpful degrees of assistance for a memory game task. These assistive levels were fine-tuned to accommodate the moves of simulated 'imperfect' users. The second layer, namely the mentalising layer, was responsible for inferring participants' strategies from previous interactions using a heuristic-based ToM. This enabled the robot to apply the learnt assistance from the first layer and explain why a specific action was chosen. 
Additionally, decoupling the ToM layer from the decision-making algorithm allowed us to provide reasoning behind the robot's actions, contributing to the design of more transparent interactive systems.
 
By combining simulation and a real-world study with N=56 untrained participants, we were able to address our research question. Indeed, our findings suggest that when the robot displayed ToM it was significantly more effective in its assistance (H1 and H2) and was perceived by participants to be more capable of anticipating their beliefs and intentions (H3) than when the robot lacked ToM abilities.

The architecture was tailored for a specific use case, but the separation of the two layers makes the upper layer, which learns the policy, generalisable to other assistive domains such as a robot providing motivations during physical exercises, while the mentalising layer based on a heuristic is more specific to the domain. 
Given the exploratory nature of the study, our findings should be viewed as indicative rather than conclusive. The patterns observed across different variables suggest meaningful trends, but the potential for false positives requires that these results be interpreted with caution and confirmed in future studies.

With this work, we aim to contribute to the field of HRI, specifically in the development of robotics applications that aid individuals. By anticipating users' needs and providing tailored assistance that aligns with their preferences, we hope to enhance interactions and increase acceptance, ultimately paving the way for more effective HRI.
Furthermore, our research highlights the importance of studying ToM in real-world situations, rather than in controlled settings, and using untrained participants for a more accurate understanding. Ultimately, this study represents a step forward in integrating complex cognitive concepts like ToM to create more intuitive, effective, and successful HRI applications. 

\section*{Statements and Declarations}

\textbf{Conflict of Interest}: The authors declare that they have no conflict of interest. Silvia Rossi is a member of the Editorial Board of the journal.
\\
\textbf{Compliance with Ethical Standards}: The study included a digital consent form describing the nature of the study. Individuals agreed to participate after reading and accepting the digital
informed consent.
\\
\textbf{Data Availability}: Data are available upon request.\\
\textbf{Funding}: This work was partially funded by the Italian Ministry for Universities and Research (MUR) under the complementary actions to the NRRP  ``Fit4MedRob - Fit for Medical Robotics'' Grant (PNC0000007), and the European Union - Next-GenerationEU - National Recovery and Resilience Plan (NRRP) - PRIN 2022 PNRR ``ADVISOR'' CUP N.E53D23016260001 (S. Rossi), and by the European Union's Horizon 2020 research and innovation programme under the Marie Skłodowska-Curie grant agreement No 801342 (Tecniospring INDUSTRY) and by the EU-founded project grant agreement No 101070930 (VALAWAI) (A. Andriella). 

\begin{appendices}






\end{appendices}


\bibliographystyle{spmpsci}      %
\bibliography{references.bib}

\end{document}